\documentclass[runningheads]{llncs}
\usepackage{graphicx}
\graphicspath{{Figures/}}

\begin{document}
\title{Understanding Aesthetic Evaluation using Deep Learning}
%
%
\author{Jon McCormack\inst{1}\orcidID{0000-0001-6328-5064} \and
 Andy Lomas\inst{2}\orcidID{0000-0003-2177-3021} }
%
%
\institute{SensiLab, Monash University, Caulfield East, Victoria 3145, Australia
\email{Jon.McCormack@monash.edu}\\
\url{http://sensilab.monash.edu} \and
Goldsmiths, University of London, London SE14 6NW, UK\\
\email{a.lomas@gold.ac.uk}\\
\url{https://www.gold.ac.uk/computing}}

\maketitle              
\begin{abstract}
A bottleneck in any evolutionary art system is aesthetic evaluation. Many different methods have been proposed to automate the evaluation of aesthetics, including measures of symmetry, coherence, complexity, contrast and grouping. The interactive genetic algorithm (IGA) relies on human-in-the-loop, subjective evaluation of aesthetics, but limits possibilities for large search due to user fatigue and small population sizes. In this paper we look at how recent advances in deep learning can assist in automating personal aesthetic judgement. Using a leading artist's computer art dataset, we use dimensionality reduction methods to visualise both genotype and phenotype space in order to support the exploration of new territory in any generative system. Convolutional Neural Networks trained on the user's prior aesthetic evaluations are used to suggest new possibilities similar or between known high quality genotype-phenotype mappings.  

\end{abstract}
\section{Introduction}
Artistic evolutionary search systems, such as the Interactive Genetic Algorithm (IGA) have been used by artists and researchers for decades \cite{Dawkins1986,Sims1991,Todd1991,Todd1992,McCormack1992,Bentley1999,Rowbottom1999,BentleyCorne2002,McCormack2019cc}. A key advantage of the IGA is that it substitutes formalised fitness measures for human judgement. The algorithm arose to circumvent the difficulty in developing generalised fitness measures for ``subjective'' criteria, such as personal aesthetics or taste. Hence the IGA found favour from many artists and designers, keen to exploit the powerful search and discovery capabilities offered by evolutionary algorithms, but unable to formalise their aesthetic judgement in computable form.

Over the years, the research community has proposed many new theories and measures of aesthetics, with research from both the computational aesthetics (CA) and psychology communities \cite{Johnson2019}. Despite much effort and many advances, a computable, universal aesthetic measure remains elusive -- an open problem in evolutionary music and art \cite{McCormack2005a}. One of the reasons for this is the psychological nature of aesthetic judgement and experience. In psychology, a detailed model of aesthetic appreciation and judgement has been developed by Leder and colleagues \cite{Leder2004,Leder2014}. This model describes information-processing relationships between various components that integrate into an aesthetic experience and lead to an aesthetic judgement and aesthetic emotion. The model includes perceptual aesthetic properties, such as symmetry, complexity, contrast, and grouping, but also social, cognitive, contextual and emotional components that all contribute in forming an aesthetic judgement. A key element of the revised model \cite{Leder2014} is that it recognises the influence of a person's affective state on many components and that aesthetic judgement and aesthetic emotion co-direct each other.

One of the consequences of this model is that any full computational aesthetic measure must take into account the affective state of the observer/participant, in addition to other factors such as previous experience, viewing context and deliberate (as opposed to automatic) formulations regarding cognitive mastering, evaluation and social discourse. All factors that are very difficult or impossible for current computational models to adequately accommodate.

How then can we progress human-computer collaboration that involves making aesthetic judgements if fully developing a machine-implementable model remains illusive? One possible answer lies in teaching the machine both tacit and learnt knowledge about an individual's personal aesthetic preferences so that the machine can assist a person in creative discovery. The machine provides assistance only, it does not assume total responsibility for aesthetic evaluation.

In this paper we investigate the use of a number of popular machine learning methods to assist digital artists in searching the large parameter spaces of modern generative art systems. The aim is for the computer to learn about an individual artist's aesthetic preferences and to use that knowledge to assist them in finding more appropriate phenotypes. ``Appropriate'' in the sense that they fit the artist's conception of high aesthetic value, or that they are in some category that is significant to the artist's creative exploration of a design space. Additionally, we explore methods to assist artists in understanding complex search spaces and use that information to explore ``new and undiscovered territory''. Finally, we discuss ways that mapping and learning about both genotype and phenotype space can inspire a search for new phenotypes ``in between'' known examples. These approaches aim to eliminate the user fatigue of traditional IGA approaches. 

\subsection{Related Work}
\label{ss:relatedWork}
In recent years, a variety of deep learning methods have been integrated into evolutionary art systems. Blair \cite{Blair2019} used adversarial co-evolution, evolving images using a GP-like system alongside a LeNet-style Neural Network critic. 

Bontrager and colleagues \cite{Botranger2018} describe an evolutionary system that uses a Generative Adversarial Network (GAN), putting the latent input vector to a trained GAN under evolutionary control, allowing the evolution of high quality 2D images in a target domain. 

Singh et al. \cite{Singh2019} used the feature vector classifier from a Convolutional Neural Network (CNN) to perform rapid visual similarity search. Their application was for design inspiration by rapidly searching for  images with visually similar features from a target image, acquired via a smartphone camera. The basis of our method is similar in the use of using a large, pre-trained network classifier (such as ResNet-50) to find visual similarity between generated phenotype images and a database of examples, however our classifier is re-trained on artist-specific datasets, increasing its accuracy in automating personal aesthetic judgement.

\section{Exploring Space in Generative Design}
\label{s:exploringSpace}

\begin{figure}
\includegraphics[width=\textwidth]{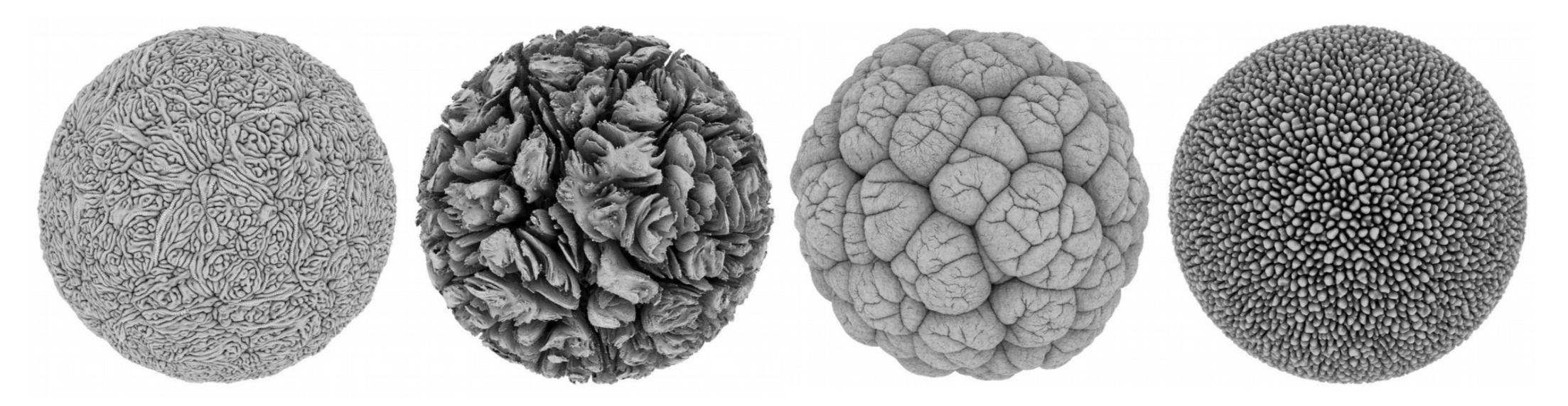}
\caption{Example cellular forms generated by cellular morphogenesis} \label{f:exampleForms}
\end{figure}

In the experiments described in this paper, we worked with a dataset of evolutionary art created by award-winning artist Andy Lomas. Lomas works with developmental morphogenetic models that grow and develop via a cellular development process. Details of the technical mechanisms of his system can be found in \cite{Lomas2014}. A vector of just 12 real valued parameters determines the resultant form, which grows from a single cell into complex forms often involving more than one million cells. The range of forms is quite varied, Figure \ref{f:exampleForms} shows a small selection of samples. In exploring the idea of machine learning of personal aesthetics, we wanted to work with a real, successful artistic system\footnote{Lomas is an award winning computer artist who exhibits internationally, see his website  \url{http://www.andylomas.com}}, rather than an invented one, as this allows us to understand the ecological validity \cite{Brunswik1956} of any system or technique developed. Ecological validity requires the assessment of creative systems in the typical environments and contexts under which they are actually experienced, as opposed to a laboratory or artificially constructed setting. It is considered an important methodology for validating research in the creative and performing arts \cite{Jausovec2011}.

\subsection{Generative Art Dataset}
\label{ss:dataset}

The dataset used consisted of 1,774 images, each generated by the developmental form generation system using the software \emph{Species Explorer} \cite{Lomas2016}. Each image is a two-dimensional rendering of a three-dimensional form that has been algorithmically grown based on 12 numeric parameters (the ``genotype''). The applied genotype determines the final developed 3D form (the ``phenotype''), which is rendered by the system to create a 2D image of the 3D form.

The dataset also contains a numeric aesthetic ranking of each form (ranging from 0 to 10, with 1 the lowest and 10 the highest, 0 meaning a failure case where the generative system terminated without generating a form). These rankings were all performed by Lomas, so represent his personal aesthetic preferences. Ranking visual form in this manner is an integral part of using his Species Explorer software, with the values in the dataset created over several weeks as he iteratively generated small populations of forms, ranked them, then used those rankings to influence the generation of the next set of forms.

Lomas has also developed a series of stylistic categorisations that loosely describe the visual class that each form fits into. This categorisation becomes useful for finding forms \emph{between} categories, discussed later. Category labels included ``brain'' (245 images), ``mess'' (466 images), ``balloon'' (138 images), ``animal'' (52 images), ``worms'' (32 images) and ``no growth'' (146 images). As each form develops through a computational growth simulation process, some of the images fail to generate much at all, leading to images that are ``empty'' (all black or all white). There were 252 empty images, leaving 1,522 images of actual forms. Even though the empty images are not visually interesting, they still hold interesting data as their generation parameters result in non-viable forms. Most of the category data (1,423 images) had been created at the same time as when Lomas was working on the original Cellular Forms. The remaining 351 images were categorised by Lomas as part of this project.

\subsection{Understanding the Design Space}
\label{ss:designSpace}

\begin{figure}
\includegraphics[width=\textwidth]{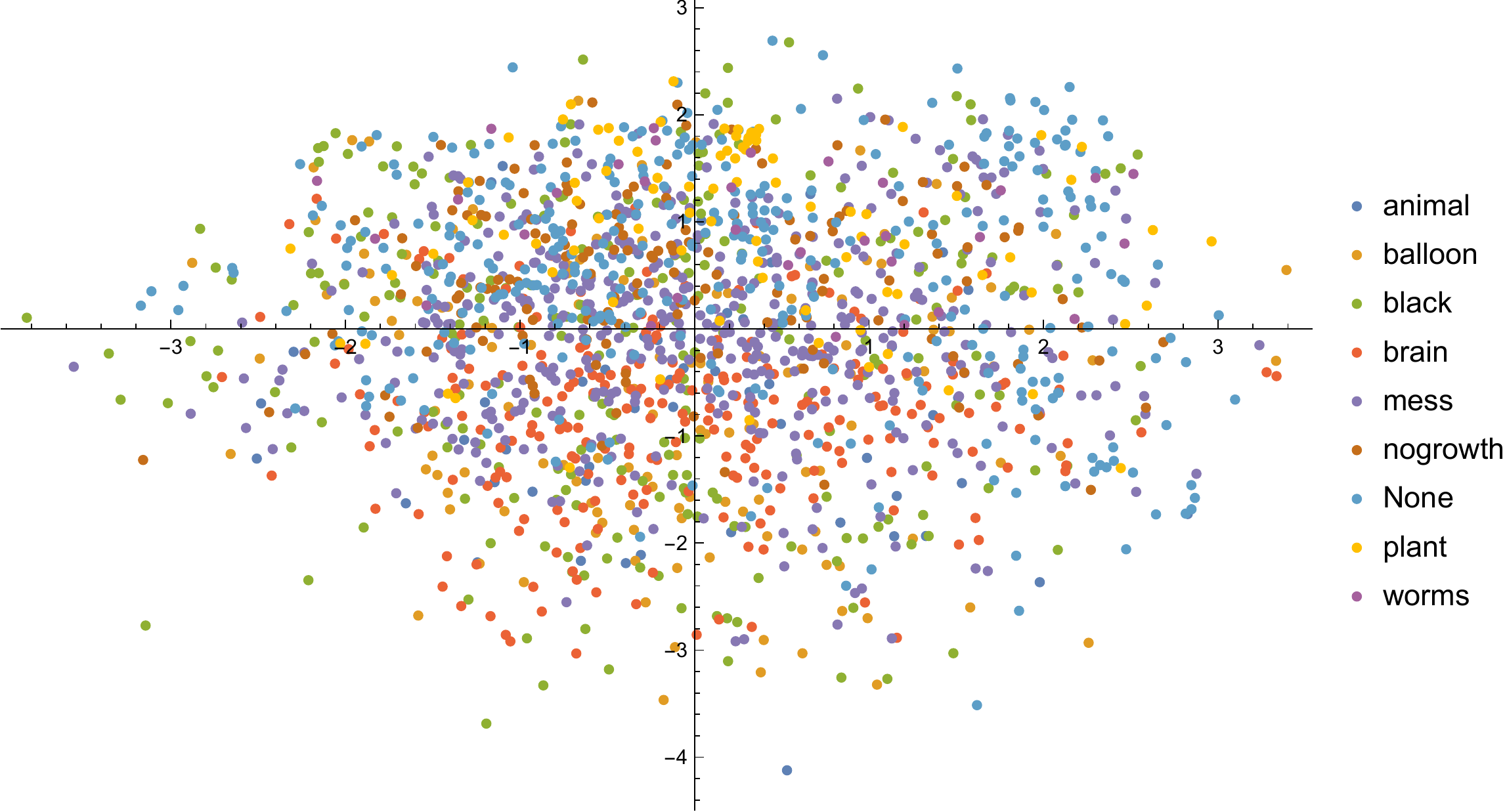}
\includegraphics[width=\textwidth]{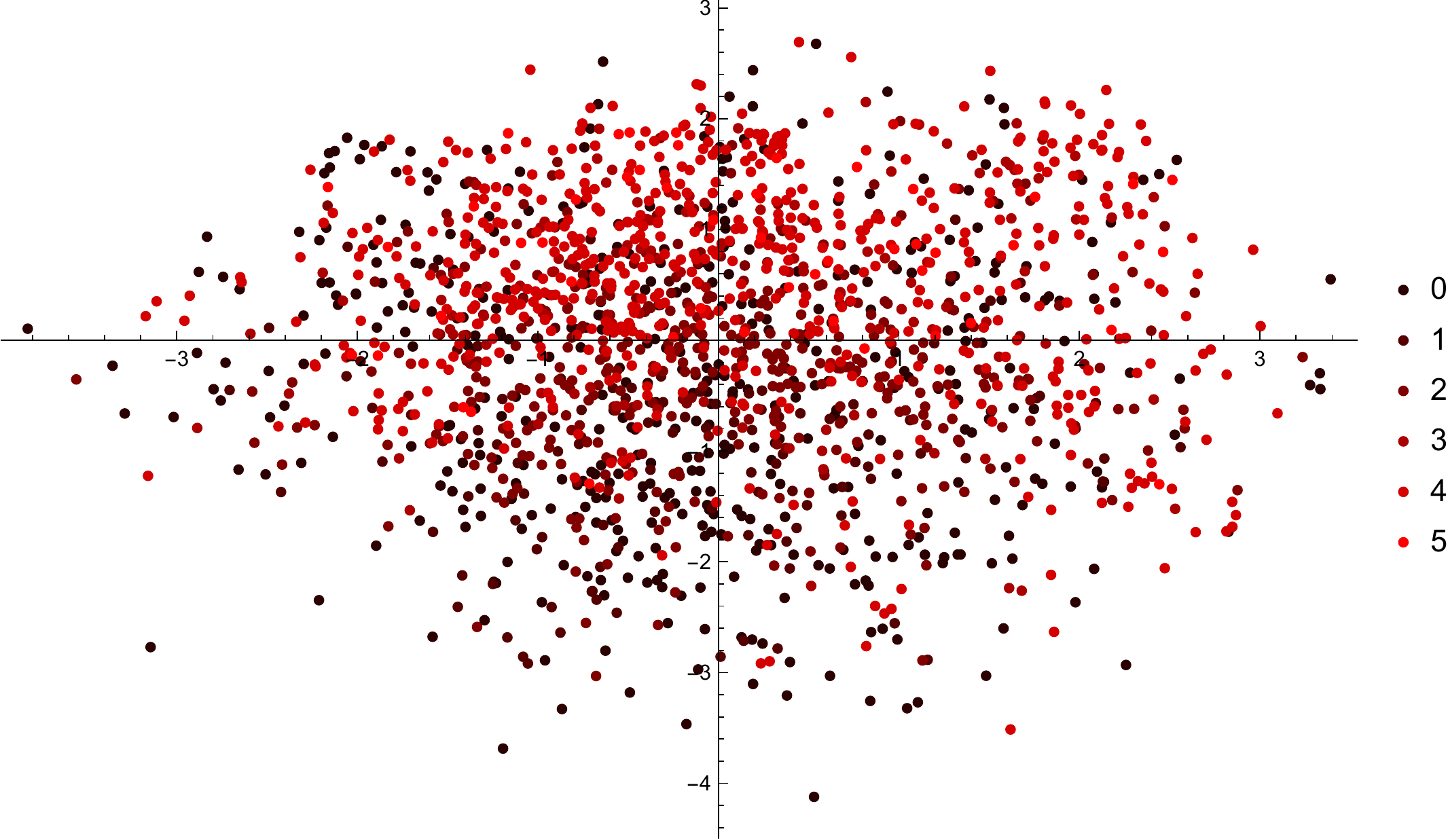}
\caption{Plot of genotype distribution in 2 dimensions using t-SNE. Individual genotypes are coloured by category (top) and by score (bottom). } \label{f:geneCat}
\end{figure}

As a first step in understanding the design space we used a variety of dimension reduction algorithms to visualise the distribution of both genotype and phenotype space to see if there was any visible clustering related to either aesthetic ranking scores or categories. We experimented with a number of different algorithms, including t-SNE \cite{maaten2008visualizing}, UMAP \cite{mcinnes2018umap} and Variational Autoencoders \cite{makhzani2015adversarial}, to see if such dimension reduction visualisation techniques could help artists better understand relationships between genotype and categories or highly ranked species.

As shown in Figure \ref{f:geneCat}, the dimensionally reduced genotype space tends to have little visible structure. The figure shows each 12-dimensional genotype dimensionally reduced to two dimensions and colour-coded according to category (top) and rating (bottom). In the case of rating, we reduced the ten-point numeric scale to five bands for clarity. The figure shows the results obtained with the t-SNE dimension reduction, testing with other algorithms (PCA, UMAP and a Variational Autoencoder) did not result in significantly better visual results.

Although some grouping can be seen in the figure, any obvious overall clustering is difficult to observe, particularly for the categories. While there is some overall structure in the score visualisation (high ranked individuals tend to concentrated in the around the upper left quadrant), discerning any regions of high or low quality is difficult. In many cases, low and high ranked individuals map to close proximity in the 2D representation.

What this analysis reveals is that the genotype space is highly unstructured in relation to aesthetic concerns, making it difficult to easily evolve high quality phenotypes. The developmental nature of the generative system, which depends on physical simulation, means that small parameter changes at critical points can result in large differences in the resultant developed form.

\subsection{Phenotype Space}
\label{ss:phenotypeSpace}
\begin{figure}
\includegraphics[width=\textwidth]{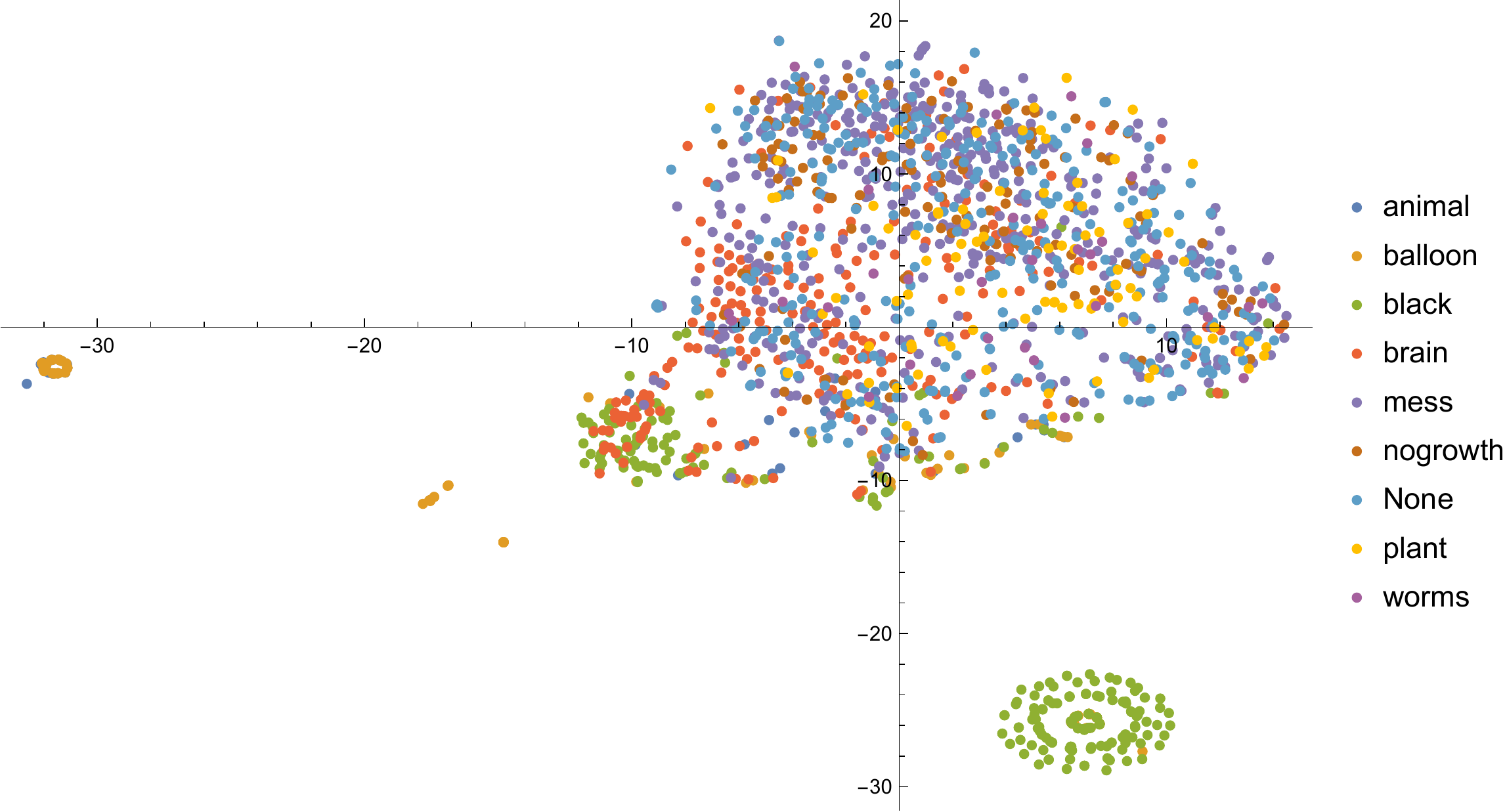}
\includegraphics[width=\textwidth]{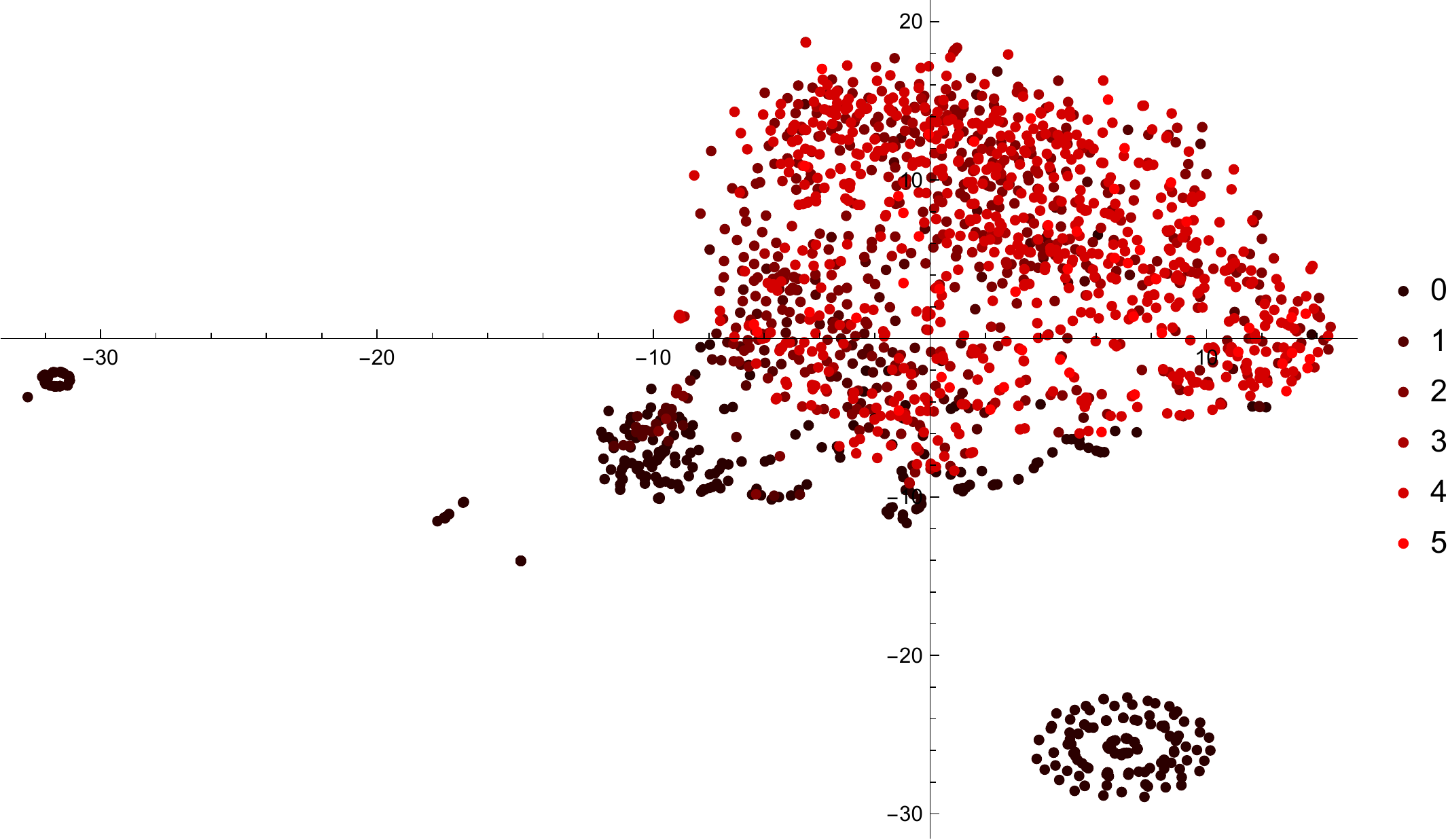}
\caption{Plot of phenotype distribution in 2 dimensions using t-SNE. Individual phenotypes are coloured by category (top) and by score (bottom). } \label{f:phenoCat}
\end{figure}

To visualise the phenotype space we used the feature classification layer of the ResNet-50 convolutional neural network. Because ResNet was trained on 1.2 million images from the ImageNet dataset \cite{deng2009imagenet}, it is very good at identifying image features that humans also recognise.  Networks trained on the ImageNet classification tasks have been shown to work very well as off the shelf image extractors \cite{sharif2014cnn}, and show even better results when fine-tuned to datasets for the task at hand \cite{azizpour2015generic}. The network produces a 2048-element vector based on the features of an input image. This vector is then dimensionally reduced to create a two-dimensional visualisation of the feature space. Again, we used the t-SNE algorithm to reduce the dimensionality of the space.

Figure \ref{f:phenoCat} shows the results for both the category (top) and score (bottom) classifications. As the figure shows, this time structure can be seen in the feature data. Classifications such as ``black'' and ``balloon'' are visible in specific regions. Similarly, the score distribution shows increasing values towards the upper-right quadrant in the visualisation.

Such visualisations can therefore potentially assist artists in navigating and understanding the space of possibilities of their generative system, because they allow them to direct search in specific regions of phenotype space. A caveat here is that the dimension reduction process ideally needs to be reversible, i.e. that one can go from low dimensions to higher if selection specific regions on a 2D plot. As a minimum, it is possible to determine a cluster of nearby phenotypes in 2D space and seed the search with the genotypes that created them, employing methods such as hill climbing to search for phenotypes with similar features.

\begin{figure}[tb]
\begin{center}
\includegraphics[width=0.7\textwidth]{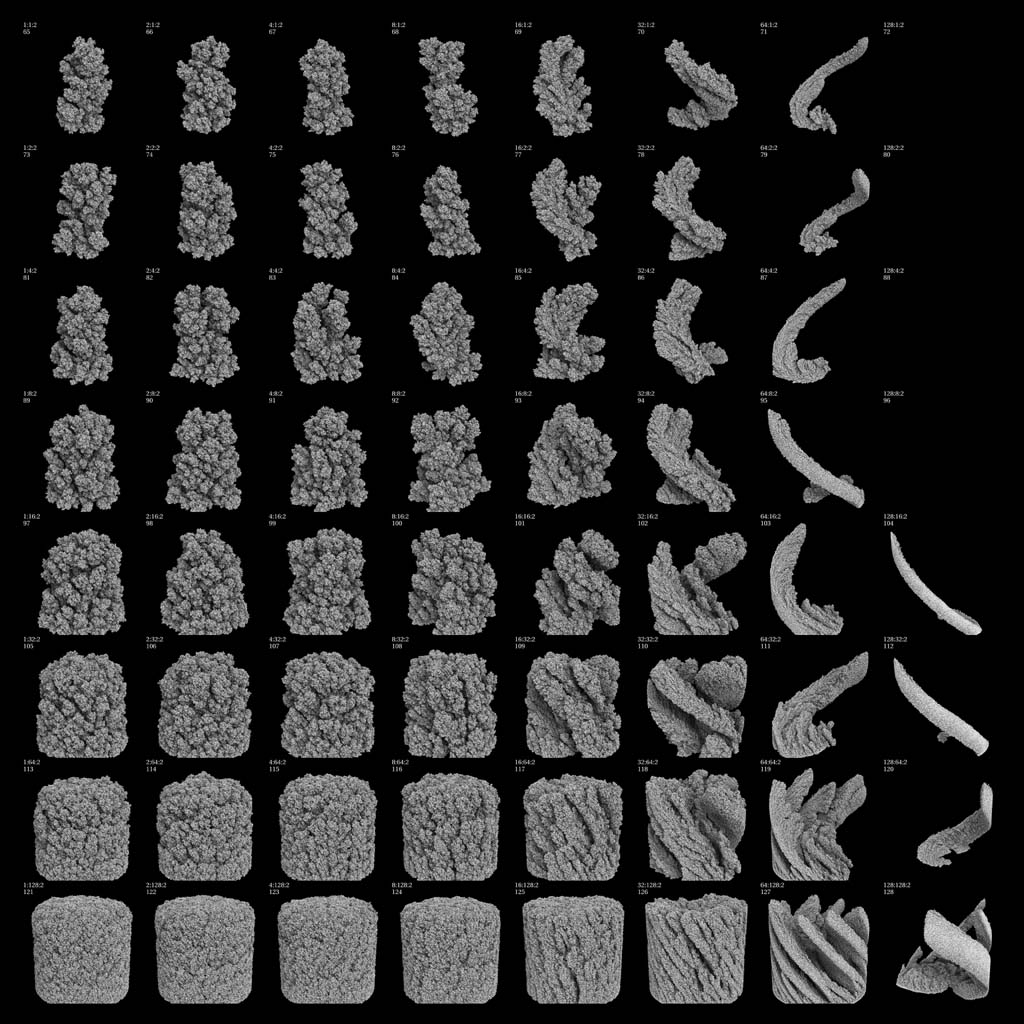}
\end{center}
\caption{Plot from Aggregation series showing effects of varying genotype parameters} \label{f:aggregationPlot}
\end{figure}

\subsection{Parameter Searching and Interpolation}
\label{ss:parameterSearch}

In early work, such as Lomas' Aggregation \cite{Lomas2005} and Flow \cite{Lomas2007} series, the artist would create plots showing how the phenotype changes depending on parameter values of the genotype. An example of such a plot can be seen in Figure \ref{f:aggregationPlot}. In these systems the genotype had a very low number of dimensions, typically just two or three parameters, which allowed a dense sampling of the space of possibilities by simply independently varying each parameter in the genotype over a specified range, running the generative system with each set of parameter values, and plotting the results in a chart with positions for each image based on the genotype parameters. One intuition from these plots is that the most interesting rich and complex behaviour often happens at transition points in the genotype space, where one type of characteristic behaviour changes into another. This can be seen in Figure \ref{f:aggregationPlot} where the forms in the 6th and 7th columns are particularly richly structured. These changes occur at parameter settings where the generative system was at a transition state between stability (to the left) and instability (to the right).

As the number of dimensions increases performing a dense sampling of the genotype space runs into the ``Curse of Dimensionality'' \cite{Bellman1961,Donoho2000}, where the number of samples needed increases exponentially with the number of parameters. Even if enough samples can be taken, how to visualise and understand the space becomes difficult and concepts such as finding the nearest neighbours to any point in the parameter space become increasingly meaningless \cite{Marimont1979}. One potential approach to make sense of higher dimensional spaces is to categorise different phenotypes created by the system. By defining categories for phenotypes we can express searching for transition points in a meaningful way as being the places in genotype space where small changes in the genotype result in changing from one phenotype category to another.

\section{Learning an Artist's Aesthetic Preferences}
\label{s:learningPreferences}

A ResNet-50 classifier was tested with with the same dataset of 1,774 images with ratings and categories as described above, using 1,421 images in the training dataset and 353 images in the validation dataset. Re-training the final classifier layers of ResNet-50 created a network that matched Lomas' original categories in the validation set with an accuracy of 87.0\%. We also looked at the confidence levels for the predictions, based on the difference between the network's probability value for the predicted category and the probability level of the highest alternative category.

\begin{table}
\begin{tabular}{ |p{6cm}|p{6cm}|  }
 \hline
Confidence quartile & Prediction accuracy\\
 \hline
75\% to 100\% & 97.1\%\\
50\% to 75\% & 97.9\%\\
25\% to 50\% & 90.5\%\\
0\% to 25\% & 67.6\%\\
 \hline
\end{tabular}
\caption{\label{tab:accuracyConfidence}ResNet-50 accuracy levels for different confidence quartiles.}
\end{table}

Table \ref{tab:accuracyConfidence} shows how the prediction accuracy varies depending on the confidence levels. The network has a reliability of over 97\% for the images in the top two confidence quartiles, with 69\% of the incorrect categorisations being in the lowest confidence quartile. A visual inspection of images in the lowest confidence quartile confirmed that these were typically also less clear which category an image should be put in to a human observer.

\begin{figure}
\begin{center}
\includegraphics[width=0.5\textwidth]{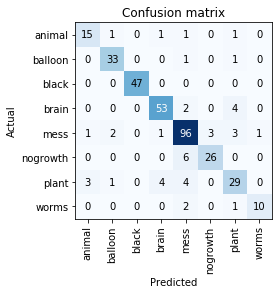}
\end{center}
\caption{Confusion matrix for ResNet-50 (phenotype space) categorisor} \label{f:resnetConfusionMatrix}
\end{figure}

The confusion matrix in Figure \ref{f:resnetConfusionMatrix} shows that the predictions appear to be consistently good across all categories, with the majority of items in each category predicted correctly. The most confused categories were ``mess'' and ``nogrowth'', both of which indicate forms that are considered by the artist to be aesthetic failure cases and sometimes look quite similar.

As well as being used for categorisation, a ResNet-50 network with a scalar output was trained against the aesthetic ranking using values from 0 to 10 that Lomas had given the forms. This resulted in a network that predicted the ranking of images in the validation set with a root mean square error of 0.716. Given that these ranking are subjective evaluations of images that often have very similar appearance this appears to be a high level of predictive accuracy.

\subsection{Genotype Space}
\label{ss:genotypeSpace}

The dataset was tested to see whether predictions of the phenotype category and aesthetic rank could be obtained from genotype parameters. This is desirable as good predictions of phenotype from the genotype values could directly aid exploration of the space of possibilities. Techniques such as Monte Carlo methods could be used to choose new candidate points in genotype space with specified fitness criteria. We could use the predictions to generate plots of expected behaviour as genotype parameters are varied that could help visualise the phenotype landscape and indicate places in genotype space where transitions between phenotype classes may occur. If meaningful gradients can be calculated from predictions, gradient descent could be used to directly navigate towards places in genotype space were one category is predicted to change into another and transitional forms between categories may exist.

Fast.ai \cite{howard2018fastai}, a Python machine learning library to create deep learning neural networks, was used to create neural net predictors for the category and aesthetic rank using genotype values as the input variables. The fast.ai Tabular model was used, with a configuration of two fully connected hidden layers of size 200 and 100. The same training and validation sets were used as previously.

Using these neural nets we achieved an accuracy of 68.3\% for predictions of the category, and predictions of the aesthetic rank had a root mean square error of 1.88. These are lower quality predictions that we obtained with the ResNet-50 classifier using the phenotype, but this is to be expected given that the ranking and categorisation are intended to be evaluations of the phenotype and were done by Lomas looking at the images of the phenotype forms. The results are also confirmed by the dimensionally-reduced visualisations presented in Section \ref{ss:designSpace}.

\begin{figure}
\begin{center}
\includegraphics[width=0.5\textwidth]{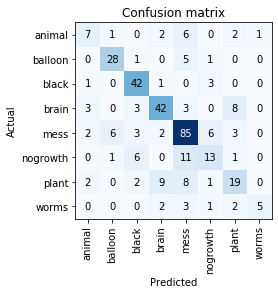}
\end{center}
\caption{Confusion matrix for Tabular (genotype space) categoriser} \label{f:tabularConfusionMatrix}
\end{figure}

The confusion matrix for the category predictions is shown in Figure \ref{f:tabularConfusionMatrix}. Similarly to the results with the ResNet-50 (phenotype space) categoriser, the ``mess'' and ``nogrowth'' categories are often confused, but with the genotype space categoriser the ``plant'' and ``brain'' categories are also quite frequently confused with each other. This suggests that it might be worth generating more training data for the ``brain'' and ``plant'' categories to improve predictive accuracy, but could also be an indication that the ``plant'' and ``brain'' categories are closely connected in genotype space.

The current version of Lomas' ``Species Explorer'' software uses a simple k-Nearest Neighbours (k-NN) method to give predictions of phenotype based on genotype data \cite{Lomas2014}. Testing with the same validation set, the k-NN method predicts categories with an accuracy of 49.8\%, and the aesthetic rank with a mean square error of 2.78. The genotype space neural net predictors give significantly better predictions than the k-NN predictor.

\begin{figure}
\includegraphics[width=\textwidth]{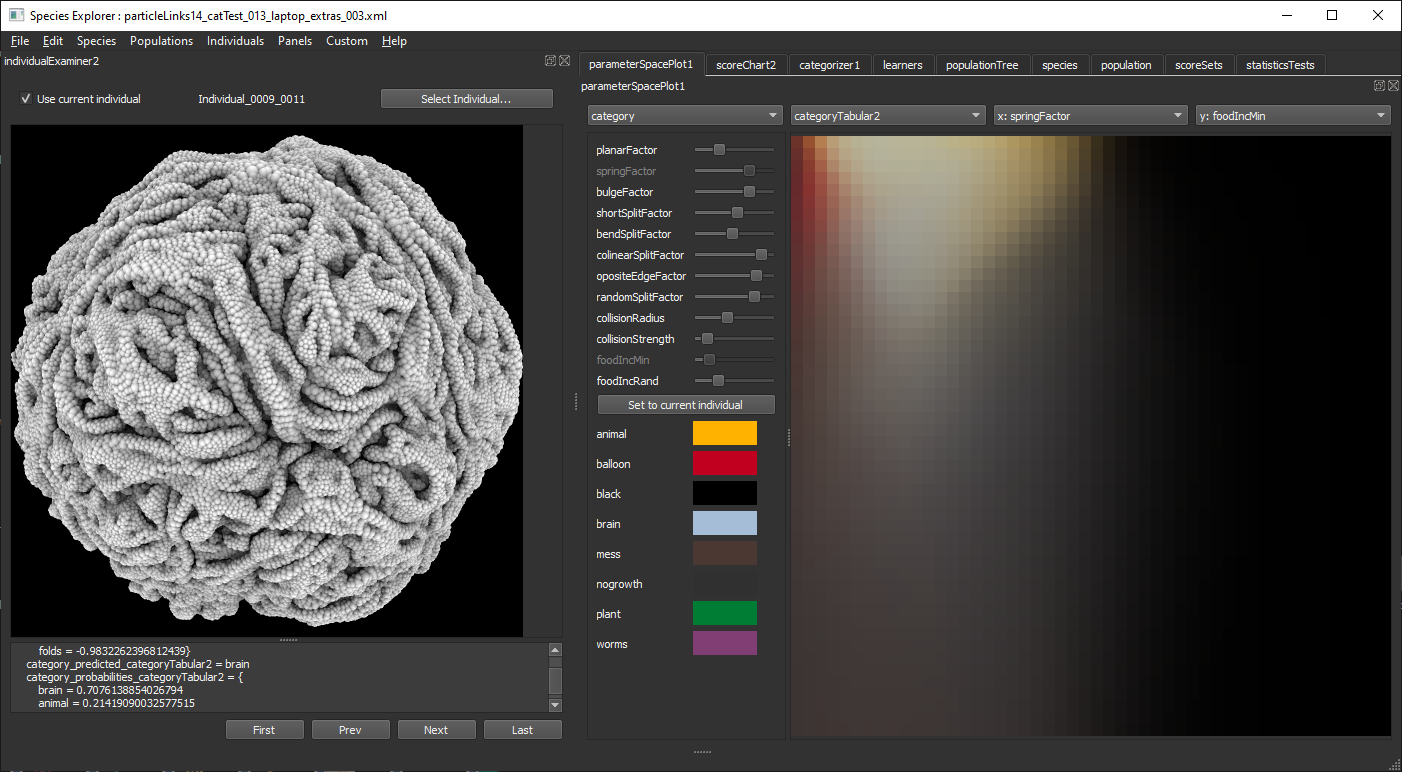}
\caption{Species Explorer user interface, showing 2D cross-section plots through genotype space using a neural net to predict the phenotype category at new positions in genotype space} \label{f:speciesExplorerGenotypeSpacePlot}
\end{figure}

A new feature was added into Species Explorer that uses the genotype neural network predictors to generate 2D cross-section plots through genotype space, showing the predicted categories or rank at different coordinate positions, see Figure \ref{f:speciesExplorerGenotypeSpacePlot}. As can be seen, these plots predict a number of potential places where transitions between categories may occur, which could lead the artist to explore new regions of the genotype space.

\section{Discussion}
\label{ss:discussion}

Our results of incorporating this new search feature into an artist's creative workflow indicate that deep learning based neural nets appear to be able to achieve good levels of accuracy when predicting the phenotype categories and aesthetic rank evaluations made by Lomas in the test dataset. The best predictions were achieved with ResNet-50, a pre-trained convolutional neural network designed for image recognition, using phenotype image data as the input. Additionally, we achieved potentially useful levels of prediction from genotype data using the fast.ai library's Tabular model to create a deep learning neural net with two fully connected hidden layers. Predictions based on the genotype rather than the phenotype are particularly interesting as they should allow navigation directly in genotype space to suggest new points to sample.

One of the main reasons for using IGAs is that the fitness function is unknown, or may not even be well defined because the artist's judgement changes over time. The use of the neural networks in this work can be seen as trying to discover whether there is a function that matches the artist's aesthetic evaluations with a useful level of predictability. If such a function can be found it could be used in a number of ways, such to use monte carlo sampling along with providing a fitness function for conventional evolutionary algorithms. If the discovered fitness function is sufficiently simple (such as being unimodal) methods like hill climbing may be appropriate.

Lomas has been using a k-Nearest Network in Species Explorer to give prediction based on position in genotype space. As the numbers of dimensions increase k-NN performance generally becomes significantly less effective \cite{Marimont1979}, while deep neural networks can still be effective predictors with higher dimensional inputs. This means the deep neural networks have the potential to allow useful levels of prediction from genotype space in systems with high numbers of genotype parameters.

It is likely that with more training data we will be able to improve the predictions from genotype space. This raises the possibility for a hybrid system: if we have a convolutional neural network that can achieve high levels of accuracy from phenotype data we could use this to automate creation of new training data for a genotype space predictor. In this way, improving the ability of a genotype space based predictor may be at least partially automated.

There is a lot of scope for trying out different configurations of deep neural networks for genotype space predictions. The choice of a network in this study, with two fully connected hidden layers with 200 and 100 neurons, was simply based on the default values suggested in the fast.ai documentation of their Tabular model. A hyper-parameter search would likely reveal even better results.

An important part of this process is how to make ranking and categorisation as easy for a creative practitioner as possible. The aim should be to allow the artist to suggest new categories and ways of ranking with as few training examples as are necessary to get good levels of prediction. It should also facilitate experimentation, making the process of trying out new ways of ranking and different ways of categorising behaviour as simple as possible.

In both authors' experience, it is often only after working with a generative system for some time, typically creating hundreds of samples, that categories of phenotype behaviour start to become apparent. This means that manually categorising all the samples that have already been generated can become significantly laborious. This has meant that although Lomas' Species Explorer software allows phenotype samples to be put into arbitrary categories, and data from categorisation can be used to change fitness functions used to generate new samples, for the majority of systems Lomas has created he hasn't divided phenotype results into categories and has relied on aesthetic rank scores instead. This is one area where pre-trained network classifiers, such as ResNet-50, may be useful. If we can reliably train a neural network to classify different phenotypes with only a small amount of training data it could make the process of creating and testing different ways of categorising phenotypes significantly easier.

\begin{figure}
\includegraphics[width=\textwidth]{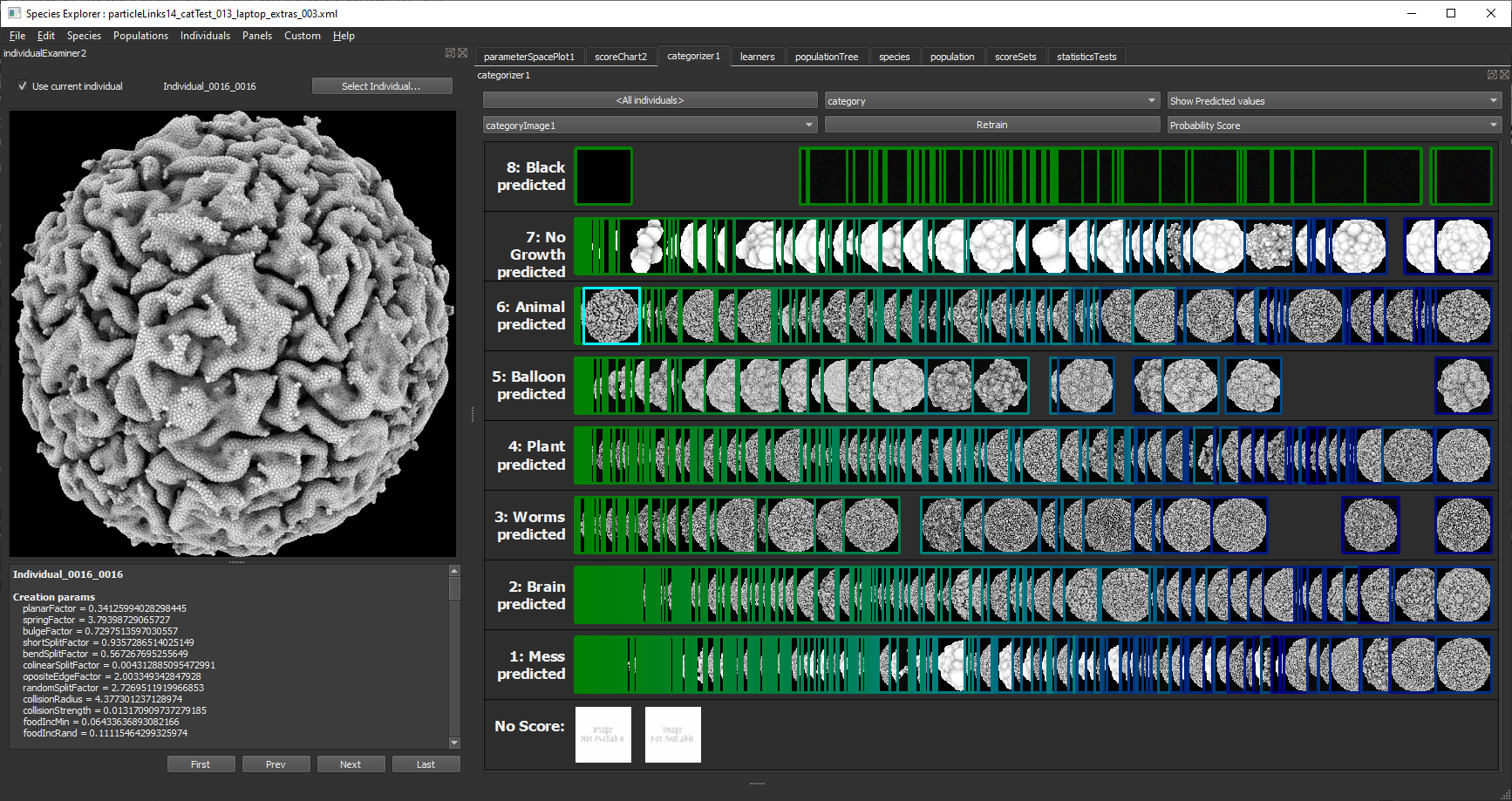}
\caption{Species Explorer user interface, showing predicted categorisations from a ResNet-50 network. The items are ordered based on the confidence levels for the predicted category, with the highest confidence level predictions to the left of each group} \label{f:speciesExplorerProbabilityScores}
\end{figure}

We modified the existing user interface in Species Explorer so that predictions of how a classifier would divide data into classes can be shown, together with placement and colouring of the outlines of thumbnails based on the confidence levels of predictions, see Figure \ref{f:speciesExplorerProbabilityScores}. This allows a simple evaluation of the quality of prediction, and helps indicate samples that might be good to add to the training set (such as incorrect predictions that the classifier has done with high confidence) to improve the quality of predictions.

The tests with dimensionally reduced plots in phenotype space using t-SNE on the feature vectors of ResNet-50 appear to show meaningful structure which may be useful to help divide samples into categories. In particular, this technique may be useful both to help initial categorisation, broadly dividing samples in phenotype space into categories, and to help sub-divide existing categories that the user wants to explore separating into different classes. The use of plots such as these may actively help experimentation, allowing the creative users to modify existing classification schemes and quickly try out different ideas of how to categorise phenotypes.

\section{Conclusions and Future Work}
\label{ss:conclusion}

The aim of this research was to progress machine-assisted aesthetic judgement based on an artist's personal aesthetic preferences. We worked with an established artist's work to give ecological validity to our system. While the results are specific to an individual artist, it is worth emphasising that the methods discussed generalise to any multi-parameter generative system whose phenotypes can be expressed as 2D images. Indeed, the Species Explorer software separates the creative evolution process from the actual generative system, allowing Species Explorer to work with \emph{any} parameter based generative system. 

The research presented here shows that deep learning neural networks can be useful to predict aesthetically driven evaluations and assist artists to find phenotypes of personally high aesthetic value. As discussed, these predictors are useful to help explore the outputs of generative systems directly in genotype space. 

There is still more research to be done however.
More testing is now needed to see how productive this is in practice when working with systems that often have high dimensional parameter spaces. We have shown the neural networks can categorise and rank phenotypes with a high accuracy in a specific instance, the next step would be to see if this approach generalises to other artists and their personal aesthetics.

\section{Acknowledgements}
This research was supported by an Australian Research Council grant FT170100033 and a Monash University International Research Visitors Collaborative Seed Fund grant.

%
%
\bibliographystyle{splncs04}
\bibliography{refs}

\begin{thebibliography}{10}
\providecommand{\url}[1]{\texttt{#1}}
\providecommand{\urlprefix}{URL }
\providecommand{\doi}[1]{https://doi.org/#1}

\bibitem{azizpour2015generic}
Azizpour, H., Sharif~Razavian, A., Sullivan, J., Maki, A., Carlsson, S.: From
  generic to specific deep representations for visual recognition. In:
  Proceedings of the IEEE conference on computer vision and pattern recognition
  workshops. pp. 36--45 (2015)

\bibitem{Bellman1961}
Bellman, R.E.: Adaptive control processes: a guided tour. Princeton university
  press (1961)

\bibitem{Bentley1999}
Bentley, P.J.: Evolutionary design by computers. Morgan Kaufmann Publishers,
  San Francisco, Calif. (1999)

\bibitem{BentleyCorne2002}
Bentley, P.J., Corne, D.W. (eds.): Creative Evolutionary Systems. Academic
  Press, London (2002)

\bibitem{Blair2019}
Blair, A.: Adversarial evolution and deep learning -- how does an artist play
  with our visual system? In: Ek{\'a}rt, A., Liapis, A., Castro~Pena, M.L.
  (eds.) Computational Intelligence in Music, Sound, Art and Design. pp.
  18--34. Springer International Publishing, Cham (2019)

\bibitem{Botranger2018}
Bontrager, P., Lin, W., Togelius, J., Risi, S.: Deep interactive evolution. In:
  Liapis, A., Romero~Cardalda, J.J., Ek{\'a}rt, A. (eds.) Computational
  Intelligence in Music, Sound, Art and Design. pp. 267--282. Springer
  International Publishing, Cham (2018)

\bibitem{Brunswik1956}
Brunswik, E.: Perception and the representative design of psychological
  experiments. University of California Press, Berkley and Los Angeles, CA, 2nd
  edn. (1956)

\bibitem{Dawkins1986}
Dawkins, R.: The Blind Watchmaker. No.~332, Longman Scientific \& Technical,
  Essex, UK (1986)

\bibitem{deng2009imagenet}
Deng, J., Dong, W., Socher, R., Li, L.J., Li, K., Fei-Fei, L.: Imagenet: A
  large-scale hierarchical image database. In: Computer Vision and Pattern
  Recognition, 2009. CVPR 2009. IEEE Conference on. pp. 248--255. Ieee (2009)

\bibitem{Donoho2000}
Donoho, D.L., et~al.: High-dimensional data analysis: The curses and blessings
  of dimensionality. AMS math challenges lecture  \textbf{1}(2000), ~32 (2000)

\bibitem{howard2018fastai}
Howard, J., et~al.: fastai. \url{https://github.com/fastai/fastai} (2018)

\bibitem{Jausovec2011}
Jausovec, N., Jausovec, K.: Brain, creativity and education. The Open
  Educational Journal  \textbf{4},  50--57 (2011)

\bibitem{Johnson2019}
Johnson, C.G., McCormack, J., Santos, I., Romero, J.: Understanding aesthetics
  and fitness measures in evolutionary art systems. Complexity
  \textbf{2019}(Article ID 3495962),  14 pages (2019).
  \doi{https://doi.org/10.1155/2019/3495962},
  \url{https://doi.org/10.1155/2019/3495962}

\bibitem{Leder2004}
Leder, H., Belke, B., Oeberst, A., Augustin, D.: A model of aesthetic
  appreciation and aesthetic judgments. British Journal of Psychology
  \textbf{95},  489--508 (2004)

\bibitem{Leder2014}
Leder, H., Nadal, M.: Ten years of a model of aesthetic appreciation and
  aesthetic judgments: The aesthetic episode – developments and challenges in
  empirical aesthetics. British Journal of Psychology  \textbf{105},  443--464
  (2014)

\bibitem{Lomas2007}
Lomas, A.: Flow, \url{http://www.andylomas.com/flow.html}

\bibitem{Lomas2005}
Lomas, A.: Aggregation: complexity out of simplicity. In: ACM SIGGRAPH 2005
  Sketches. p.~98. ACM (2005)

\bibitem{Lomas2014}
Lomas, A.: Cellular forms: An artistic exploration of morphogenesis. In: AISB
  2014 - 50th Annual Convention of the AISB (07 2014)

\bibitem{Lomas2016}
Lomas, A.: Species explorer: An interface for artistic exploration of
  multi-dimensional parameter spaces. In: Bowen, J., Lambert, N., Diprose, G.
  (eds.) Electronic Visualisation and the Arts (EVA 2016). Electronic Workshops
  in Computing (eWiC), BCS Learning and Development Ltd., London (12th-14th
  July 2016)

\bibitem{maaten2008visualizing}
Maaten, L.v.d., Hinton, G.: Visualizing data using t-sne. Journal of machine
  learning research  \textbf{9}(Nov),  2579--2605 (2008)

\bibitem{makhzani2015adversarial}
Makhzani, A., Shlens, J., Jaitly, N., Goodfellow, I., Frey, B.: Adversarial
  autoencoders. arXiv preprint arXiv:1511.05644  (2015)

\bibitem{Marimont1979}
Marimont, R., Shapiro, M.: Nearest neighbour searches and the curse of
  dimensionality. IMA Journal of Applied Mathematics  \textbf{24}(1),  59--70
  (1979)

\bibitem{McCormack1992}
McCormack, J.: Interactive evolution of forms. In: Cavallaro, A., Harley, R.,
  Wallace, L., Wark, M. (eds.) Cultural Diversity in the Global Village: Third
  International Symposium on Electronic Art. p.~122. The Australian Network for
  Art and Technology, Sydney, Australia (1992)

\bibitem{McCormack2005a}
McCormack, J.: Open problems in evolutionary music and art. In: Rothlauf, F.,
  Branke, J., Cagnoni, S., Corne, D.W., Drechsler, R., Jin, Y., Machado, P.,
  Marchiori, E., Romero, J., Smith, G.D., Squillero, G. (eds.) EvoWorkshops.
  Lecture Notes in Computer Science, vol.~3449, pp. 428--436. Springer (2005)

\bibitem{McCormack2019cc}
McCormack, J.: Creative systems: A biological perspective. In: Veale, T.,
  Cardoso, F.A. (eds.) Computational Creativity: The Philosophy and Engineering
  of Autonomously Creative Systems, pp. 327--352. Springer Nature, Switzerland,
  AG (2019). \doi{10.1007/978-3-319-43610-4}

\bibitem{mcinnes2018umap}
McInnes, L., Healy, J., Melville, J.: {UMAP}: Uniform manifold approximation
  and projection for dimension reduction. arXiv preprint arXiv:1802.03426
  (2018)

\bibitem{Rowbottom1999}
Rowbottom, A.: Evolutionary Design by Computers, chap. Evolutionary Art and
  Form, pp. 261--277. Morgan Kaufmann, San Francisco, CA (1999)

\bibitem{sharif2014cnn}
Sharif~Razavian, A., Azizpour, H., Sullivan, J., Carlsson, S.: Cnn features
  off-the-shelf: an astounding baseline for recognition. In: Proceedings of the
  IEEE conference on computer vision and pattern recognition workshops. pp.
  806--813 (2014)

\bibitem{Sims1991}
Sims, K.: Artificial evolution for computer graphics. In: Computer Graphics.
  vol.~25, pp. 319--328. ACM SIGGRAPH, ACM SIGGRAPH, New York (July 1991),
  \url{http://www.genarts.com/karl/papers/siggraph91.html}

\bibitem{Singh2019}
Singh, D., Rajic, N., Colton, S., McCormack, J.: Camera obscurer: Generative
  art for design inspiration. In: Evolutionary and Biologically Inspired Music,
  Sound, Art and Design - 8th International Conference, EvoMUSART 2019.
  Springer-Nature, Berlin (April 2019)

\bibitem{Todd1991}
Todd, S., Latham, W.: Mutator: a subjective human interface for evolution of
  computer sculptures. Tech. rep. (1991)

\bibitem{Todd1992}
Todd, S., Latham, W.: Evolutionary Art and Computers. Academic Press, London
  (1992)

\end{thebibliography}

\end{document}